\documentclass[10pt,twocolumn,letterpaper]{article}

\usepackage{cvpr}
\usepackage{times}
\usepackage{epsfig}
\usepackage{graphicx}
\usepackage{amsmath}
\usepackage{amssymb}
\usepackage{color}
\usepackage{dsfont}
\usepackage[lined,ruled,commentsnumbered]{algorithm2e}
\usepackage{enumitem}

\cvprfinalcopy 

\def\cvprPaperID{****} 

\begin{document}

\title{Explainable Deep Classification Models for Domain Generalization}

	\author{Andrea Zunino$^*$\textsuperscript{,}$^\dagger$\textsuperscript{,1}, Sarah Adel Bargal\thanks{Equal contribution.}  \textsuperscript{ ,}\textsuperscript{2}, Riccardo Volpi\thanks{This work was done while A. Zunino and R. Volpi were at Istituto Italiano di Tecnologia.} \textsuperscript{ ,}\textsuperscript{3},  Mehrnoosh Sameki\textsuperscript{4},\\ Jianming Zhang\textsuperscript{5}, Stan Sclaroff\textsuperscript{2}, Vittorio Murino\textsuperscript{1,6,7}, Kate Saenko\textsuperscript{2}\\~\\ 
		\textsuperscript{1}Huawei Ireland Research Center 
		\textsuperscript{2}Department of Computer Science, Boston University \\ 
		\textsuperscript{3}Naver Labs Europe
		\textsuperscript{4}Microsoft
		\textsuperscript{5}Adobe Research \\
		\textsuperscript{6}Pattern Analysis \& Computer Vision, Istituto Italiano di Tecnologia
		\textsuperscript{7}Universit\`a di Verona\\
		{\tt\small \{andrea.zunino,riccardo.volpi,vittorio.murino\}@iit.it,}\\ 
		{\tt\small \{sbargal,sclaroff,saenko\}@bu.edu, mehrnoosh.sameki@microsoft.com, jianmzha@adobe.com}}

  

%

\maketitle

\begin{abstract}

Conventionally, AI models are thought to trade off explainability for lower accuracy. We develop a training strategy that not only leads to a more explainable AI system for object classification, but as a consequence, suffers no perceptible accuracy degradation. Explanations are defined as regions of visual evidence upon which a deep classification network makes a decision. This is represented in the form of a saliency map conveying how much each pixel contributed to the network's decision. Our training strategy enforces a periodic saliency-based feedback to encourage the model to focus on the image regions that directly correspond to the ground-truth object. We quantify explainability using an automated metric, and using human judgement. We propose explainability as a means for bridging the visual-semantic  gap  between  different  domains where model explanations are used as a means of disentagling domain specific information from otherwise relevant features. We demonstrate that this leads to improved generalization to new domains without hindering performance on the original domain. 


\end{abstract}

\begin{figure}[!t]
    \centering
    \includegraphics[width=\linewidth,trim={0cm 0cm 0cm 0cm},clip]{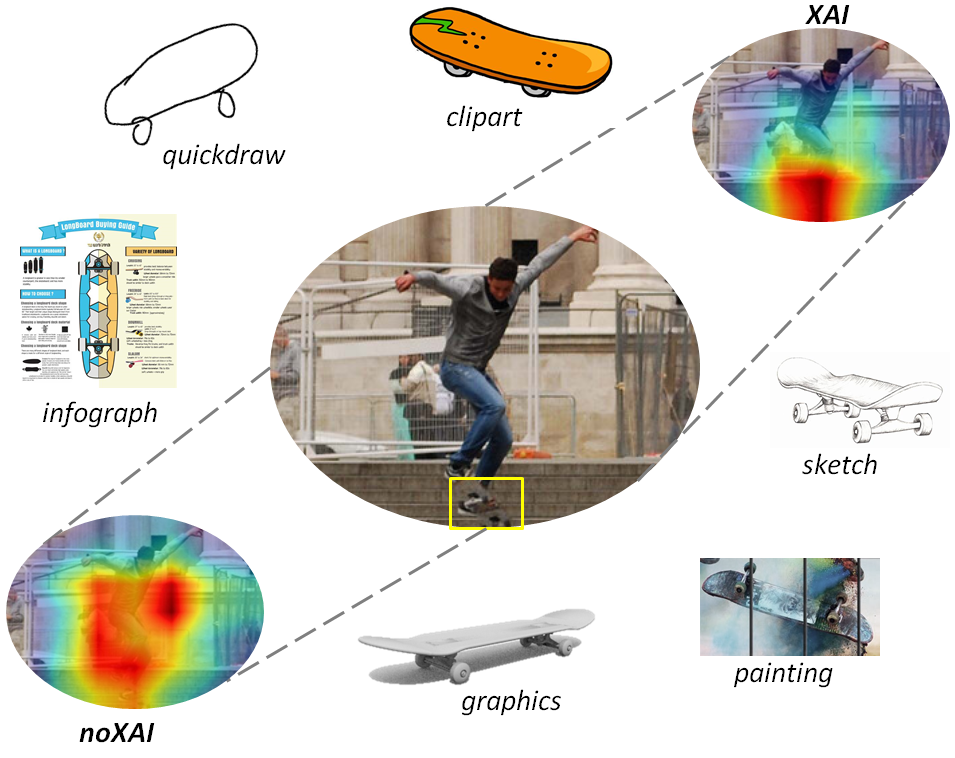}
\caption{In this figure we demonstrate how explainability (XAI) can be used to achieve domain generalization from a \textit{single source}. Training a deep neural network model to enforce explainability, \eg focusing on the skateboard region (red is most salient, and blue is least salient) for the ground-truth class skateboard in the central training image, enables improved generalization to other domains where the background is not necessarily class-informative.}
\label{fig:domains}
\end{figure}

\section{Introduction}



Increased explainability in machine learning is traditionally associated with lower performance, \eg a decision tree is more explainable, but less accurate than a deep neural network. In this paper we argue that, in fact, increasing the explainability of a deep classifier can improve its generalization, especially to novel domains. End-to-end deep models often exploit biases unique to their training dataset which leads to poor generalization on novel datasets or environments. We develop a training strategy for deep neural network models that increases explainability, suffers  no  perceptible  accuracy  degradation on the training domain, and improves performance on unseen domains.  

Domain adaptation and generalization are formulations that mitigate the problem of dataset bias. In domain adaptation one needs to know a priori the target distribution, which limits applicability \cite{Daume2006,Blitzer2006,Saenko2010}. In standard domain generalization techniques, one needs several source domains for training, which may not be available in practice. A more generic formulation is \textit{single-source} domain generalization, where one would like to avoid learning dataset bias for better generalization, but only has access to a \textit{single} source distribution. In this work, we address this challenging \textit{single-source} setting. Data augmentation approaches were shown to be successful for improving generalization to unseen domains by randomizing and perturbing the way training images are portrayed, and therefore learning invariance against some inherent biases of the source dataset~\cite{DomainRandomization,volpi2018neurips,volpi2019iccv}. 

A limitation of such data augmentation techniques is that some biases do not depend on color features, but are more structured or context dependent. Consider the case where every sample associated with a given class always has the same background, \eg an object recognition dataset where a soccer ball is mostly seen on a soccer field. In this case, the classifier might learn the background features instead of the ball characteristics -if they are sufficient to obtain good accuracy on training. While context and context-correlations aid classification when the source and target domains come from similar distributions~\cite{oliva2007role,torralbacontext,choi2010exploiting}, this becomes a limitation when the correlation is corrupted in unseen target domains. 
The drawback is evident: a model that learns to recognize soccer balls by evaluating whether grass is present or not in a scene will poorly generalize to scenarios where grass is not present or visible. This vulnerability, commonly referred to as ``domain bias'', significantly limits the applicability of machine learning systems into the wild. 

We posit that the design of algorithms that better mimic the way humans reason, or ``explain", can help mitigating the domain bias issue. Our approach utilizes explainability as a means for bridging the visual-semantic gap between different domains as presented in Figure~\ref{fig:domains}. Specifically, our training strategy is guided by model explanations and available human-labeled explanations. 
Explanations are defined as regions of visual evidence upon which a network makes a decision. This is represented in the form of a saliency map conveying how much each pixel contributed to the network's decision. 

Our training strategy periodically guides the forward activations of spatial layer(s) of a Convolutional Neural Network (CNN) trained for object classification. The activations are guided to focus on regions in the image that directly correspond to the ground-truth class label, as opposed to context that may more likely be domain dependent. The proposed strategy aims to reinforce explanations that are non-domain specific, and alleviate explanations that are domain specific. Classification models are compact and fast in comparison to more complex semantic segmentation models. Our approach allows the compact classification model to possess some properties of a segmentation model without increasing model complexity or test-time overhead.

We show how the identification of evidence within a visual input using top-down neural attention formulations \cite{gradcam} can be a powerful tool for domain analysis. Inspired by these findings, we demonstrate that more explainable deep classification models could be trained without hindering their performance. We then conduct a human study to confirm our intuitive quantification of an ``explainable" model. Finally, we demonstrate how the explainable model better generalizes on six unseen target domains, although it was trained only using a single-source domain. 



In summary, our contributions are:
\begin{itemize}
    \item We propose a training strategy that leads to more explainable deep classification models. We quantify explainability computationally and using human judgement.
    \item We demonstrate benefits of having a more explainable model for single-source domain generalization.
\end{itemize}

The rest of the paper is organized as follows. Section~\ref{rel_work} reviews related works. Section~\ref{domain_analysis} introduces how explainability can be used to analyze domain evidence and domain shift. Section~\ref{method} introduces our proposed training strategy. Section~\ref{exps} presents our experimental setup and results for quantifying explainability and domain generalization on the task of object classification. We then conclude the paper.

\section{Related Work}
\label{rel_work}

\textbf{Robustness to Domain Shift.} Several problem formulations have been proposed with the aim of learning models which are more robust in out-of-distribution settings. One formulation that received a large amount of interest form the community is domain adaptation~\cite{Daume2006,Blitzer2006,Saenko2010}. The assumption here is to have access to a set of samples from a ``source'' domain for which annotations are available, and a set of samples from a ``target'' distribution on which we desire to perform well, for which annotation is unavailable or only partially available. There is a significant body of works that propose effective solutions to this problem (\eg, ~\cite{Daume2006,Blitzer2006,Saenko2010,Ganin,ADDA,DeepCORAL}); the limitation is that one needs to know the target distribution a priori. 

In domain generalization~\cite{muandet2013icml,li2017iccv,motiian2017iccv,li2018aaai,shankar2018iclr,Li_2019_ICCV,zunino2020predicting} this assumption is relaxed; the goal here is to learn models that better generalize to unseen domains, without fixing target distributions a priori. As it was originally conceived, domain generalization requires access to several source domains to learn models that better generalize. Recently, different works have proposed ways to learn more general representations by relying on a single-source distribution. Volpi \etal~\cite{volpi2018neurips} propose to rely on adversarial robustness~\cite{goodfellow2015iclr}, generating samples that are hard for the model over iterations. A related method~\cite{volpi2019iccv} proposes to find new data augmentation rules over iterations by evaluating the image transformations that the current model is more vulnerable to. Carlucci \etal~\cite{Carlucci_2019_CVPR} rely on self-supervised learning to learn representations that are less biased towards the source distribution. Domain randomization~\cite{DomainRandomization} allows improving performance on unseen, real data when training on rendered data. Hendricks \etal~\cite{hendricks2018women} enforce looking at a person as opposed to looking at other background elements in an image to make gender prediction less biased.

In this work, we show that our proposed training procedure improves domain generalization performance of the learned models, without assumptions on the target distributions, and without the need of multiple source distributions. With respect to models trained with data augmentation strategies~\cite{volpi2019iccv,volpi2018neurips} for domain generalization, the method proposed here allows overcoming more complex dataset biases. We show that our learning procedure is complementary to data augmentation, and can be efficiently used in tandem. 

\textbf{Saliency for Explainability.} The “black-box” nature of end-to-end
deep neural networks creates highly non-linear and inexplicable feature representations that make it difficult to understand what causes the models to make certain decisions -evidence of a model prediction. Various methods have been introduced that investigate this major drawback of such powerful models. For visual data, interpretability/explainability has been addressed in the form of saliency maps highlighting image regions that a model uses to make a prediction, \ie evidence. Explanation of visual models has been addressed using white-box methods where the model parameters are assumed to be known, and using black-box methods where model parameters are assumed to be unknown.

White-box methods include ~\cite{zhang2018top,bargal2017excitation,deconv,simonyan2013deep,grad,CAM,gradcam}.
Zeiler \etal \cite{deconv} use a variant of the standard backpropagation error from neuron activations in higher layers down to the image level. Selvaraju \etal \cite{gradcam} obtain activation maps of a specific class using a weighted sum of deep convolutional features. In~\cite{zhang2018top} top-down attention of a convolutional neural network (CNN) classifier is modelled for generating task-specific attention maps. This work was then extended in the temporal dimension in~\cite{bargal2017excitation} for Recurrent Neural Networks (RNNs) to provide visual explanations of spatiotemporal models. Black-box methods include \cite{8237633,Petsiuk2018rise} where image regions are preturbed and network output is monitored to determine regions of discriminative evidence.

Several works have employed explainability in developing training time and testing time frameworks to further improve model predictions. Cao \etal \cite{cao2015look} use explanation maps to feed regions of highest importance into the same model and use the predicted class-conditional probabilities to improve the original ones corresponding to the whole image. Zunino \etal ~\cite{zunino2018excitation} propose a guided dropout regularizer for deep networks based on the explanation of a network prediction defined as the firing of neurons in specific paths. The explanation at each neuron is utilized to determine the probability of dropout, rather than dropping out neurons uniformly at random as in standard dropout. Bargal \etal \cite{bargal2019guidedzoom} employ explainability by making sure the model has ``the right reasons'' for a prediction, defined as reasons that are coherent with those used to make similar correct decisions at training time.  Selvaraju \etal \cite{selvaraju2019taking} optimize the alignment between human attention maps and gradient-based network importance for improving performance on Visual Question Answering and Image Captioning tasks. Explainability has also been used in spatial semantic segmentation tasks \cite{li2018tell,zhou2018weakly,wei2017object}, and object localization tasks \cite{zhang2018adversarial}.  In contrast to previous works, our approach proposes a training strategy to ensure that models periodically learn to rely on object-related visual concepts for the task of object classification in order to achieve more explainable models that generalize better on unseen target domains.


\begin{figure}[!t]
    \centering
    \textit{\hspace{2.3em} Image \hspace{1.9em}  Graphics Evidence \hspace{0.4em}   Real Evidence}\\[0.2em]
    {\rotatebox[origin=c]{90}{\textit{Motorcycle}\hspace*{-5em}}} \hspace{0.2em}
    \includegraphics[width=0.29\linewidth,height=0.2\linewidth,trim={0cm 0cm 0cm 0cm},clip]{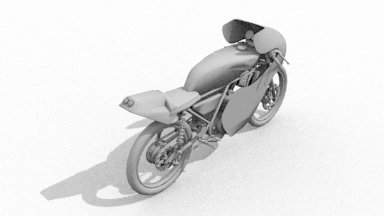}
    \hspace*{0.1em}
    \includegraphics[width=0.29\linewidth,height=0.2\linewidth,trim={0cm 0cm 0cm 0cm},clip]{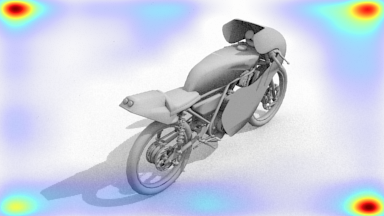}
    \hspace*{0.1em}
    \includegraphics[width=0.29\linewidth,height=0.2\linewidth,trim={0cm 0cm 0cm 0cm},clip]{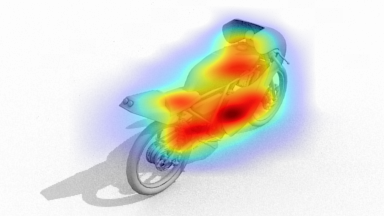} \\[0.2em] 
    {\rotatebox[origin=c]{90}{\textit{Car}\hspace*{-4em}}} \hspace{0.2em}
    \includegraphics[width=0.29\linewidth,height=0.2\linewidth,trim={0cm 0cm 0cm 0cm},clip]{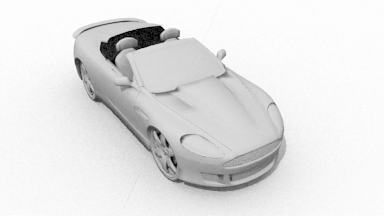}
    \hspace*{0.1em}
    \includegraphics[width=0.29\linewidth,height=0.2\linewidth,trim={0cm 0cm 0cm 0cm},clip]{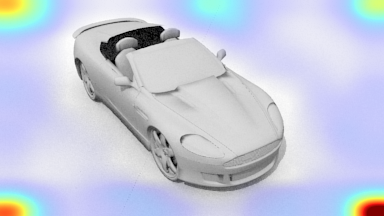}
    \hspace*{0.1em}
    \includegraphics[width=0.29\linewidth,height=0.2\linewidth,trim={0cm 0cm 0cm 0cm},clip]{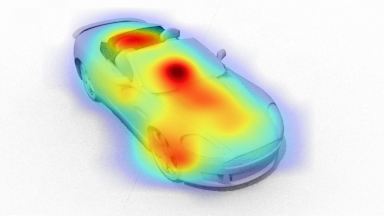} \\[0.2em]
    {\rotatebox[origin=c]{90}{\textit{Horse}\hspace*{-4em}}} \hspace{0.2em}
    \includegraphics[width=0.29\linewidth,height=0.2\linewidth,trim={0cm 0cm 0cm 0cm},clip]{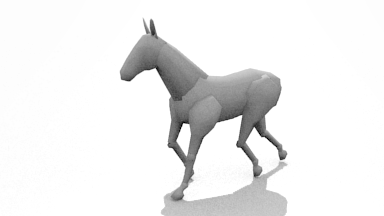}
    \hspace*{0.1em}
    \includegraphics[width=0.29\linewidth,height=0.2\linewidth,trim={0cm 0cm 0cm 0cm},clip]{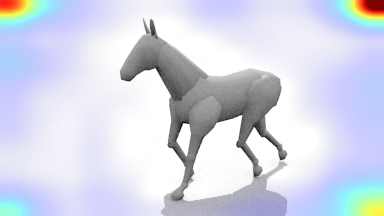}
    \hspace*{0.1em}
    \includegraphics[width=0.29\linewidth,height=0.2\linewidth,trim={0cm 0cm 0cm 0cm},clip]{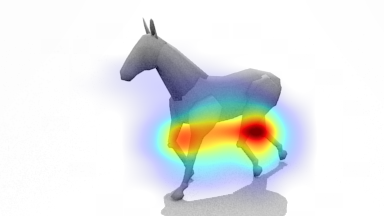} \\[0.2em]
    {\rotatebox[origin=c]{90}{\textit{Motorcycle}\hspace*{-4em}}}  \hspace{0em}
    \includegraphics[width=0.29\linewidth,height=0.2\linewidth,trim={0cm 0cm 0cm 0cm},clip]{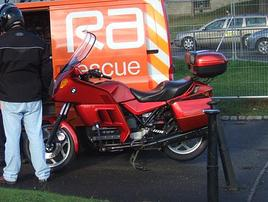}
    \hspace*{0.1em}
    \includegraphics[width=0.29\linewidth,height=0.2\linewidth,trim={0cm 0cm 0cm 0cm},clip]{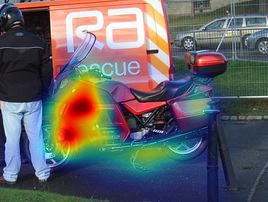}
    \hspace*{0.1em}
    \includegraphics[width=0.29\linewidth,height=0.2\linewidth,trim={0cm 0cm 0cm 0cm},clip]{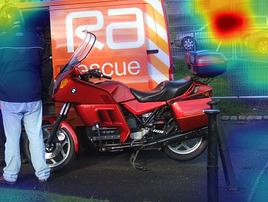} \\[0.2em]
    {\rotatebox[origin=c]{90}{\textit{Car}\hspace*{-4em}}} \hspace{0.2em}
    \includegraphics[width=0.29\linewidth,height=0.2\linewidth,trim={0cm 0cm 0cm 0cm},clip]{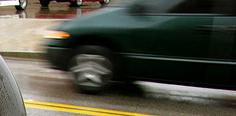}
    \hspace*{0.1em}
    \includegraphics[width=0.29\linewidth,height=0.2\linewidth,trim={0cm 0cm 0cm 0cm},clip]{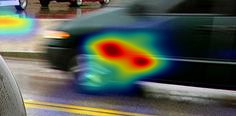}
    \hspace*{0.1em}
    \includegraphics[width=0.29\linewidth,height=0.2\linewidth,trim={0cm 0cm 0cm 0cm},clip]{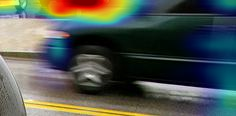} \\[0.2em]
    {\rotatebox[origin=c]{90}{\textit{Horse}\hspace*{-4em}}} \hspace{0.2em}
    \includegraphics[width=0.29\linewidth,height=0.2\linewidth,trim={0cm 0cm 0cm 0cm},clip]{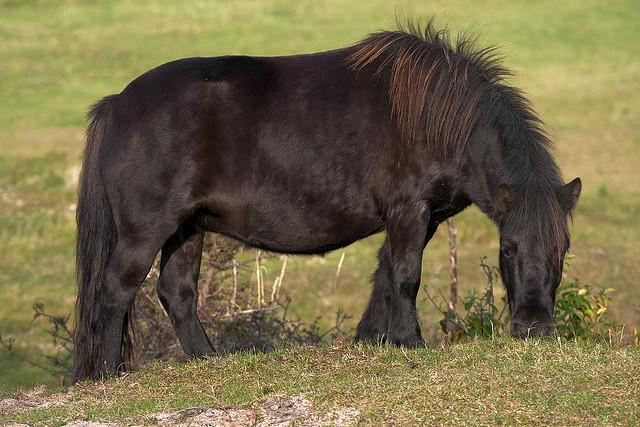}
    \hspace*{0.1em}
    \includegraphics[width=0.29\linewidth,height=0.2\linewidth,trim={0cm 0cm 0cm 0cm},clip]{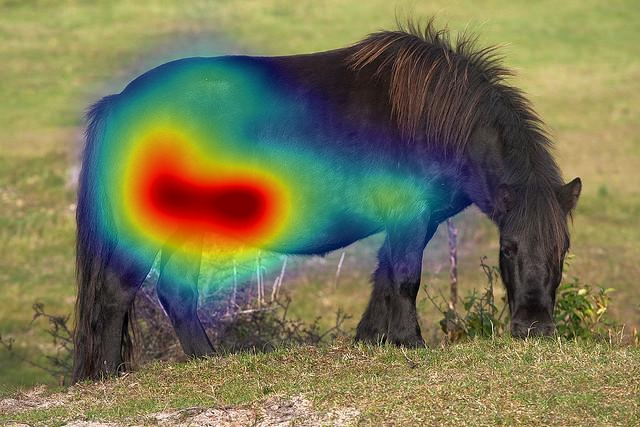}
    \hspace*{0.1em}
    \includegraphics[width=0.29\linewidth,height=0.2\linewidth,trim={0cm 0cm 0cm 0cm},clip]{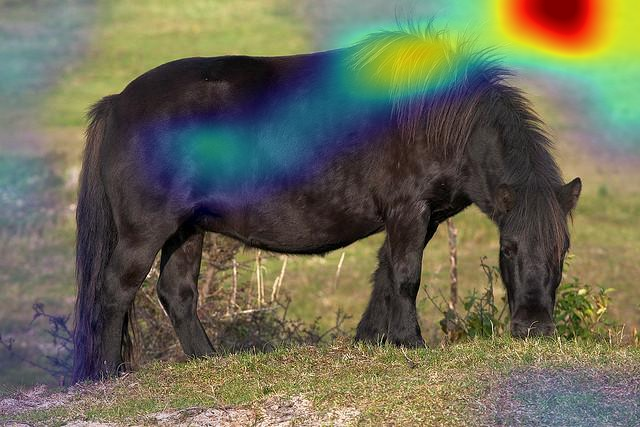} \\[0.2em]
\caption{\textbf{Domain Evidence.} 
(Left) Graphics and Real images from the classes \textit{motorcycle}, \textit{car} and \textit{horse}. (Middle) Evidence the domain discrimination network identifies for the Graphics domain. (Right) Evidence the domain discrimination network identifies for the Real domain. For Graphics images, the model selects the white background as evidence for the Graphics domain and selects the object as evidence for the Real domain. For the Real images, the model selects the objects to be evidence for the Graphics domain and selects the background as evidence for the Real domain. 
}
\label{fig:DA_2}
\end{figure}

\section{Explainability for Domain Analysis}
\label{domain_analysis}
We start by motivating how saliency can be used to highlight discriminative evidence found in each domain in a domain transfer setting. We then demonstrate how saliency can be used to highlight how different training strategies shift the model focus within the image. In this section, we use the Syn2Real  dataset \cite{peng2017visda}, which is constructed from a graphics source domain rendered from 3D CAD models and a real images target domain of the following classes: Airplane, Bicycle, Bus, Car, Horse, Knife, Motorcycle, Person, Plant, Skateboard, Train, and Truck.

\textbf{Highlighting Domain Evidence.} We set up an experiment to visualize image regions that are domain specific. We train a VGG16 network to differentiate between the graphics and real images domains of the Syn2Real dataset in a binary classification problem setting. Having a classifier trained to differentiate between domains, we can then visualize why the model processes an unseen image as belonging to a specific domain and not the other. As saliency methods can visualize evidence of classes that are not necessarily ground-truth, we visualize the evidence for each domain in images of the source domain and images of the target domain in Figure~\ref{fig:DA_2}. For graphics images, the model uses the white background as evidence for the graphics domain, and the object as evidence for the real images domain. For real images, the model uses the object as evidence for the graphics domain, and the busy background as evidence for the real images domain. The evidence associated with the ground-truth domain of an image is observed to be context-dependent, and the evidence associated with an alternative domain is observed to be object-dependent. This capability of interpreting models and visually analyzing differences between domains suggests the possibility of building models that bridge exactly that highlighted domain gap.

\begin{figure}[!t]
    \centering
    \textit{\hspace{0.5em} Image \hspace{4.5em}  no DA \hspace{4.5em}   DA}\\[0.2em]
    {\rotatebox[origin=c]{90}{\textit{Airplane}\hspace*{-4.2em}}} \hspace{0em}
    \includegraphics[width=0.29\linewidth,height=0.2\linewidth,trim={0cm 0cm 0cm 0cm},clip]{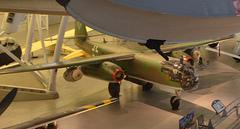} 
    \hspace*{0.1em}
    \includegraphics[width=0.29\linewidth,height=0.2\linewidth,trim={0cm 0cm 0cm 0cm},clip]{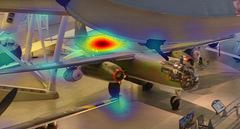} 
    \hspace*{0.1em}
    \includegraphics[width=0.29\linewidth,height=0.2\linewidth,trim={0cm 0cm 0cm 0cm},clip]{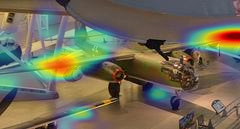}\\[0.2em]
    {\rotatebox[origin=c]{90}{\textit{Plant}\hspace*{-4em}}} \hspace{0.2em}   
    \includegraphics[width=0.29\linewidth,height=0.2\linewidth,trim={0cm 0cm 0cm 3cm},clip]{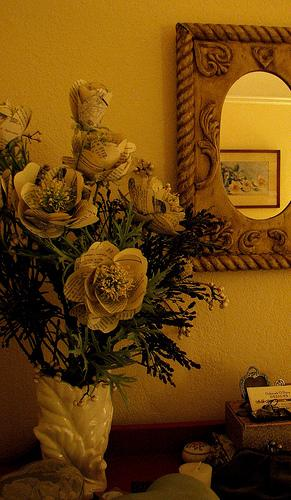} 
    \hspace*{0.1em}
    \includegraphics[width=0.29\linewidth,height=0.2\linewidth,trim={0cm 0cm 0cm 3cm},clip]{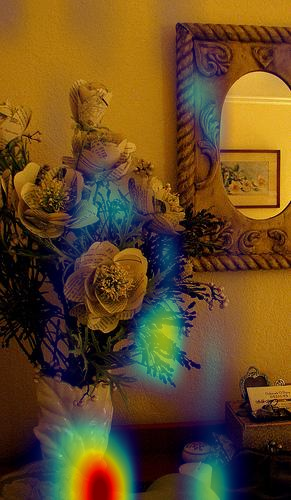} 
    \hspace*{0.1em}
    \includegraphics[width=0.29\linewidth,height=0.2\linewidth,trim={0cm 0cm 0cm 3cm},clip]{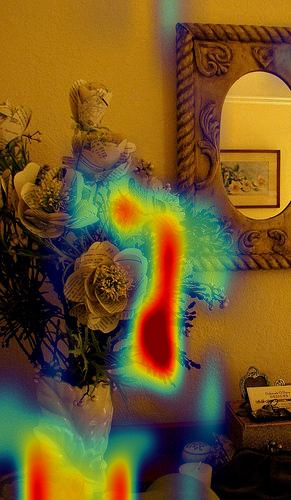}\\[0.2em]
    {\rotatebox[origin=c]{90}{\textit{Horse}\hspace*{-4em}}} \hspace{0.2em}   
    \includegraphics[width=0.29\linewidth,height=0.2\linewidth,trim={0cm 0cm 0cm 0cm},clip]{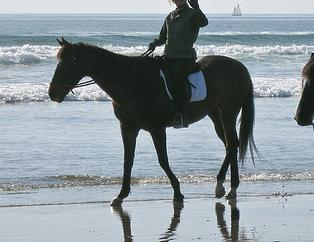} 
    \hspace*{0.1em}
    \includegraphics[width=0.29\linewidth,height=0.2\linewidth,trim={0cm 0cm 0cm 0cm},clip]{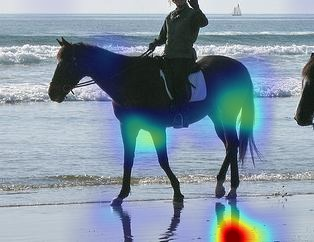} 
    \hspace*{0.1em}
    \includegraphics[width=0.29\linewidth,height=0.2\linewidth,trim={0cm 0cm 0cm 0cm},clip]{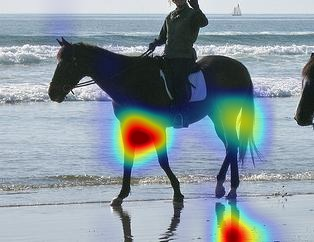}\\[0.2em]
    {\rotatebox[origin=c]{90}{\textit{Bicycle}\hspace*{-8em}}} \hspace{0em}   
    \includegraphics[width=0.29\linewidth,height=0.35\linewidth,trim={0cm 0cm 0cm 0cm},clip]{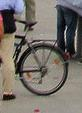} 
    \hspace*{0.1em}
    \includegraphics[width=0.29\linewidth,height=0.35\linewidth,trim={0cm 0cm 0cm 0cm},clip]{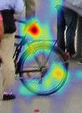} 
    \hspace*{0.1em}
    \includegraphics[width=0.29\linewidth,height=0.35\linewidth,trim={0cm 0cm 0cm 0cm},clip]{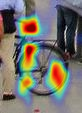}\\[0.2em]
\caption{\textbf{Domain Shift.} We train two models on the Graphics source domain of Syn2Real, with and without Domain Adaptation (DA), and then test on the Real target domain. (Left) Three sample images from the target domain of the Syn2Real dataset that were misclassified before domain adaptation, then were correctly classified after domain adaptation. (Middle) The saliency of a classification model that does not employ domain adaptation. (Right) The saliency of a classification model that employs domain adaptation. It is clear how the evidence, after domain adaptation, is more focused on discriminative evidence of the ground-truth class.}
\label{fig:DA_3}
\end{figure}

\textbf{Highlighting the Evidence Shift.} We now set up an experiment to visualize the shift of focus in input images when different training strategies are used. The first training strategy is vanilla CNN training with no domain adaptation for object classification, \ie training on graphics images only and testing on real images only. The second training strategy employs domain adaptation. We train an AlexNet on the Syn2Real classification task without Domain Adaptation. We then repeat training with the domain adaptation approach of Long \etal \cite{long2016deep}. This approach aligns distributions of the source and target domains based on a joint maximum mean discrepancy. We highlight the shift of the object's class evidence in test images of the target domain that are misclassified by the model that does not perform domain adaptation, and are correctly classified by the domain adaptation model. Examples are presented in Figure~\ref{fig:DA_3}. The saliency maps demonstrate the shift toward more discriminative evidence when the domain adaptation model is used. For example, before domain adaptation a metallic surface of the airplane was the evidence the network used to incorrectly classify the airplane as a motorcycle. However, after domain adaptation the image is correctly classified as an airplane and the evidence has shifted to the wings of the airplane - a more discriminative feature of airplane.

Testing on data from domains that are different from those used in training poses several challenges that are addressed in the fields of domain adaptation and generalization. We find that different training strategies, \eg with or without domain adaptation, make models reason based on different evidence. This inspired us to encourage models to focus on object-dependent evidence, rendering models that (1) are more explainable, and (2) improve generalization on unseen domains.

\section{Method}
\label{method}

We propose to explicitly disentangle domain specific information from otherwise relevant features using model explanations. We train models to produce saliency maps that are more explainable, in the sense that they better localize ground-truth class objects. As we will show, this results in improved performance on unseen domains. We will refer to such explainable models as Explainable AI (XAI) models, and other vanilla CNN models trained using the conventional approaches without explicitly requiring any notion of explainability as noXAI models.



At training time, we periodically (with a certain frequency $freq$) force the model to focus, within an image $x^i$, on the objects corresponding to the ground-truth label $y^i$ using the ground-truth spatial annotation $g^i$, rather than focusing on the surrounding evidence which may be more domain specific. We assume that $g^i$ is a computed 2D binary map that has 1 in a pixel location in the ground-truth object segmentation, and 0 otherwise. We also assume that $g^i$ is rescaled to correspond to the layer spatial dimension where XAI is applied to.

During a training epoch where explainability is enforced, we compute saliency maps for the ground-truth class and examine whether their peak saliency overlaps with the ground-truth spatial annotation. If the peak saliency overlaps with the ground-truth spatial annotation, we classify this saliency map to be explainable, \ie properly classified for the correct reasons. Otherwise, we enforce explainability by utilizing the ground-truth spatial annotation as an improved explanation. 

We enforce focusing on objects in an image by scaling the forward activations of a particular spatial layer $l$ in the network at certain epochs. We generate a multiplicative binary $mask$ for guiding the focus of the network in the layer in which we are enforcing XAI. For an explainable image $x^i$, the binary mask is a binarization of the achieved saliency map, \ie $mask^i_{j,k} = \mathds{1}(s^i_{j,k} > 0) ~\forall j ~\forall k$, $j=1,\dots,W$ and $k=1,\dots,H$, where $W$ and $H$ are the spatial dimension of a layers' output neuron activations; The mask is active at locations of non-zero saliency. This re-inforces the activations corresponding to the active saliency regions that have been classified as being explainable. For images that need an improved explanation, the binary mask is assigned to be the ground-truth spatial annotation $mask^i_{j,k} = g^i_{j,k} ~\forall j ~\forall k$, $j=1,\dots,W$ and $k=1,\dots,H$; The mask is active at ground-truth locations. This increases the frequency at which the network reinforces activations at locations that are likely to be non-domain specific and suppresses activations at locations that are likely to be domain specific. We then perform element-wise multiplication of our computed $mask$ with the forward activations of layer $l$; \ie $a^{l,i}_{j,k} = mask^i_{j,k} * a^{l,i}_{j,k} ~\forall j ~\forall k$, $j=1,\dots,W$ and $k=1,\dots,H$. Our XAI approach is summarized in Algorithm~\ref{alg2zoom}.

\section{Experiments}
\label{exps}

In this section, we present experimental setup and results that quantify the explainability of XAI and noXAI classification models. We  
\begin{algorithm}[h!]
	\caption{XAI Training Strategy}\label{alg2zoom}    
	\vspace{.2 cm}
	\KwIn{$x^i, i \in {1, \dots, m}$ training images; $y^i, i \in {1, \dots, m}$ corresponding class label; $g^i, i \in {1, \dots, m}$ corresponding annotation; $freq$ of training feedback; initial model $M$; $n$ epochs; layer $l$}
    \vspace{.2 cm}
	\KwOut{Trained Model $M'$}
    \vspace{.2 cm} 		 
	{\bf Procedure}:\vspace{.2 cm}\\
	\nl For every epoch $e \in 1, \dots, n$\\ 
	\nl \hspace*{1em} if $e~mod~freq~==~0$:\\ 
    \nl \hspace*{2em} For every training example $x^i, i \in 1, \dots, m$\\ 
	\nl \hspace*{3em} $s^i = saliency(x^i, y^i)$\\
	\nl \hspace*{3em} $w,h = argmax(s^i)$\\
    \nl \hspace*{3em} if $g^i_{w,h} == 1$\\ 
    \nl \hspace*{4em} // $j=1, \dots, W; k=1, \dots, H$\\ 
	\nl \hspace*{4em} $mask^i_{j,k} = \mathds{1}(s^i_{j,k} > 0) ~\forall j ~\forall k$ \\ 
    \nl \hspace*{3em} else:\\ 
	\nl \hspace*{4em} $mask^i = g^i$ \\ 
    \nl \hspace*{3em} $a^{l,i} = mask^i * a^{l,i}$ \\	
    \nl \hspace*{3em} Compute gradients and update weights\\ 
    \nl \hspace*{1em} else:\\ 
    \nl \hspace*{2em} Compute gradients and update weights
 \end{algorithm}
then present how XAI models lead to
better single-source domain generalization.

\textbf{Datasets.} We use Microsoft Common Objects in Context (MSCOCO)~\cite{lin2014microsoft} for training models using the noXAI and XAI strategies. MSCOCO provides ground-truth spatial annotations consisting of object segmentations of the corresponding ground-truth class. We use such annotations to guide the XAI training strategy. We then test how both strategies generalize from a single-source dataset to six unseen target domains from DomainNet~\cite{peng2018moment} and Syn2Real~\cite{peng2017visda}: graphics, clipart, infograph, painting, quickdraw, and sketch. To do so, we train models for the single-label classification task, once using MSCOCO as the single source, and another using the PASCAL VOC that is smaller-scale dataset with object annotations. We select the common subset of classes between source and target domains for each scenario. For MSCOCO, this leads to $\sim$25K training images 
belonging to the following classes:  Airplane, Bicycle, Bus, Car, Horse, Knife, Motorcycle, Skateboard, Train, Truck. For PASCAL VOC this leads to $\sim$11K training images belonging to the following classes: Airplane, Bicycle, Bird, Bus, Car, Cat, Chair, Cow, Dog, Horse, Motorbike, Sheep, Television, Train.

\textbf{Experimental Setup.}  
For all experiments considering noXAI, XAI we focus on a vanilla Resnet-50 architecture. We resize input images to be 224x224 pixels, and train for 50 epochs using a learning rate of $0.00001$. Saliency maps are computed using the GradCAM \cite{gradcam} algorithm after the last block layer $l$ of the ResNet-50. We choose the last spatial layer since it performed best, as it models higher level spatial patterns. In all experiments we set the frequency of XAI training to be five epochs. We observe similar results for $freq = 10,15$. We compare the performance of our models against the data augmentation strategy by Volpi and Murino~\cite{volpi2019iccv}, implemented following the recipe proposed by the authors (random concatenation of five different transformations among \textit{sharpness, brightness, color, contrast, RGB-to-grayscale conversion, RGB-channel perturbations} applied to each batch during training).


\subsection{More Explainable Classification Models}
In this section, we compare the explainablity of noXAI \vs XAI classification models using an intuitive automated metric from the computer vision community, and then using human judgment in a crowdsourcing setting.

\subsubsection{Automated Metric}
\label{automated_metric}

\begin{figure}[!t]
    \centering
    \textit{\hspace{1em} Image \hspace{4.2em}  no XAI \hspace{4.8em}  XAI}\\[0.2em]
    {\rotatebox[origin=c]{90}{\textit{Airplane}\hspace*{-4.2em}}} \hspace{0em}
    \includegraphics[width=0.29\linewidth,height=0.2\linewidth,trim={0cm 0cm 0cm 0cm},clip]{imgs/localization_img1.jpg} 
    \hspace*{0.1em}
    \includegraphics[width=0.29\linewidth,height=0.2\linewidth,trim={0cm 0cm 0cm 0cm},clip]{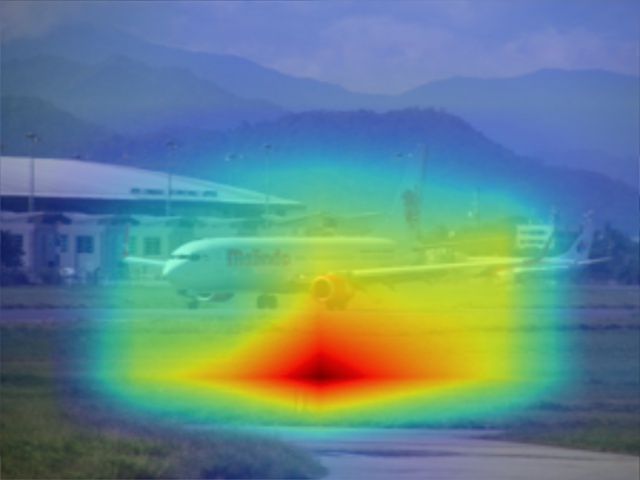} 
    \hspace*{0.1em}
    \includegraphics[width=0.29\linewidth,height=0.2\linewidth,trim={0cm 0cm 0cm 0cm},clip]{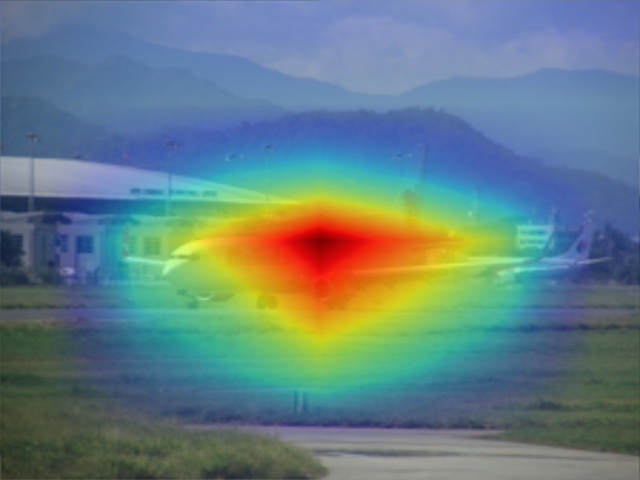}\\[0.2em]
    {\rotatebox[origin=c]{90}{\textit{Bus}\hspace*{-4.2em}}} \hspace{0.22em}
    \includegraphics[width=0.29\linewidth,height=0.2\linewidth,trim={0cm 0cm 0cm 0cm},clip]{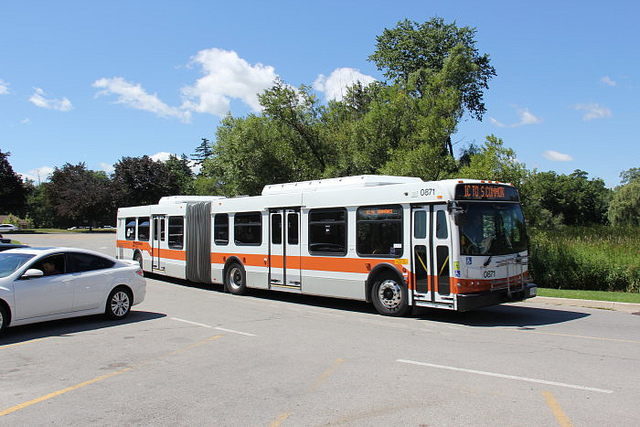} 
    \hspace*{0.1em}
    \includegraphics[width=0.29\linewidth,height=0.2\linewidth,trim={0cm 0cm 0cm 0cm},clip]{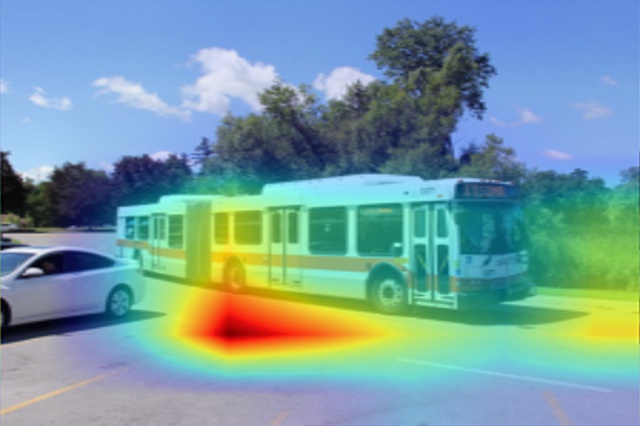} 
    \hspace*{0.1em}
    \includegraphics[width=0.29\linewidth,height=0.2\linewidth,trim={0cm 0cm 0cm 0cm},clip]{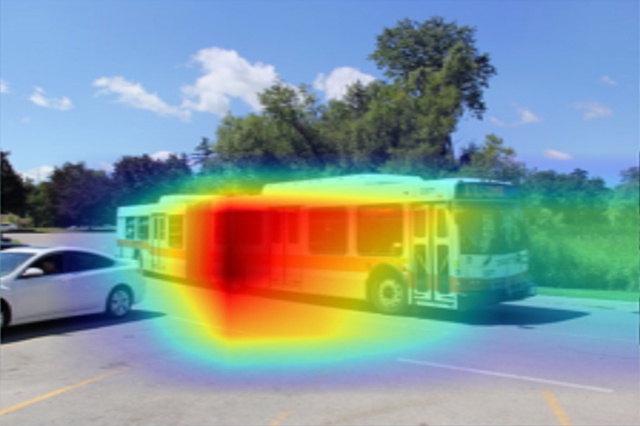}\\[0.2em]
       {\rotatebox[origin=c]{90}{\textit{Horse}\hspace*{-4.2em}}} \hspace{0.22em}
    \includegraphics[width=0.29\linewidth,height=0.2\linewidth,trim={0cm 0cm 0cm 0cm},clip]{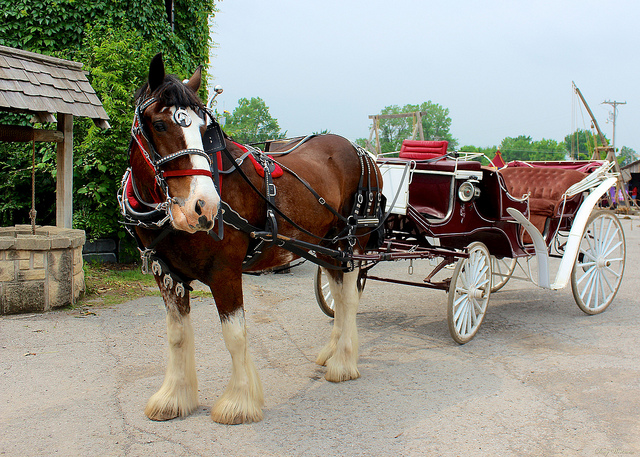} 
    \hspace*{0.1em}
    \includegraphics[width=0.29\linewidth,height=0.2\linewidth,trim={0cm 0cm 0cm 0cm},clip]{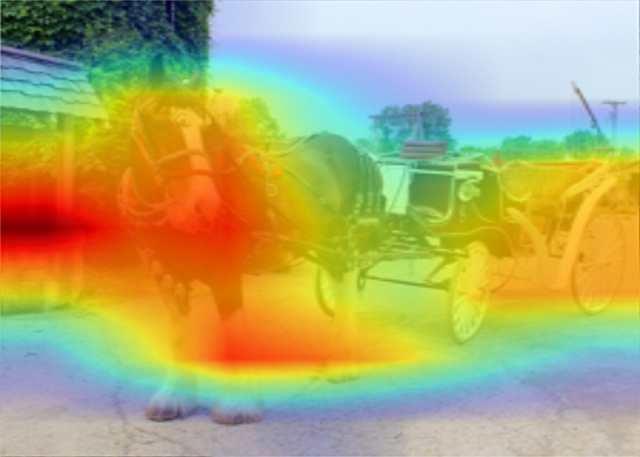} 
    \hspace*{0.1em}
    \includegraphics[width=0.29\linewidth,height=0.2\linewidth,trim={0cm 0cm 0cm 0cm},clip]{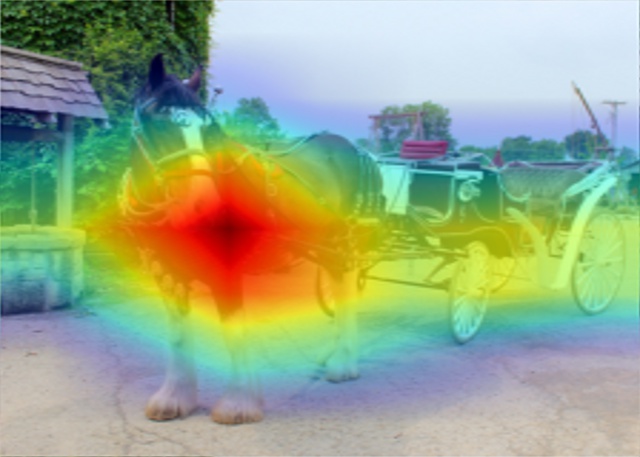}\\[0.2em]
           {\rotatebox[origin=c]{90}{\textit{Skateboard}\hspace*{-4em}}} \hspace{0.22em}
    \includegraphics[width=0.29\linewidth,height=0.2\linewidth,trim={0cm 0cm 0cm 0cm},clip]{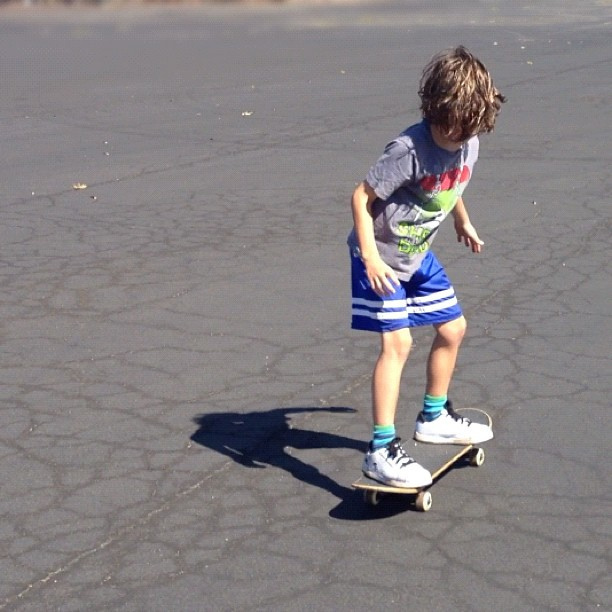} 
    \hspace*{0.1em}
    \includegraphics[width=0.29\linewidth,height=0.2\linewidth,trim={0cm 0cm 0cm 0cm},clip]{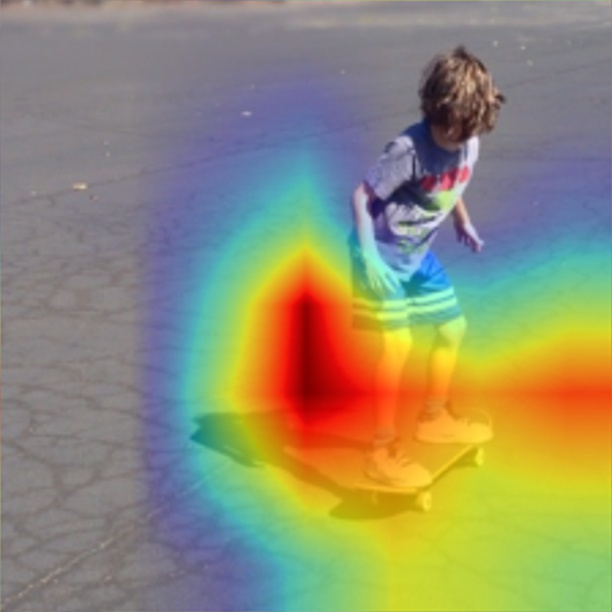} 
    \hspace*{0.1em}
    \includegraphics[width=0.29\linewidth,height=0.2\linewidth,trim={0cm 0cm 0cm 0cm},clip]{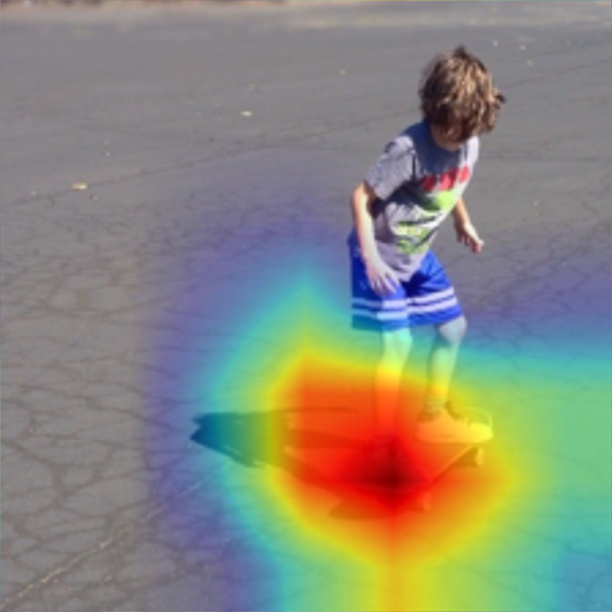}\\[0.3em]
\caption{(Left) Four sample images from the unseen MSCOCO validation set that are correctly classified by both the noXAI and XAI models. (Middle) Saliency associated with the correctly predicted class using the noXAI model. (Right) saliency associated with the correctly predicted class using the XAI model. The XAI model, based on human spatial annotations, provides feedback that enables saliency to be better localized over the objects corresponding to the ground-truth class compared to the noXAI vanilla training of a deep model, for unseen validation data. }
\label{fig:noXAI_XAI}
\end{figure}

\begin{figure}[!t]
    \centering
    \includegraphics[width=1\linewidth, height=0.75\linewidth,trim={0.5cm 0cm 1.5cm 0.65cm},clip]{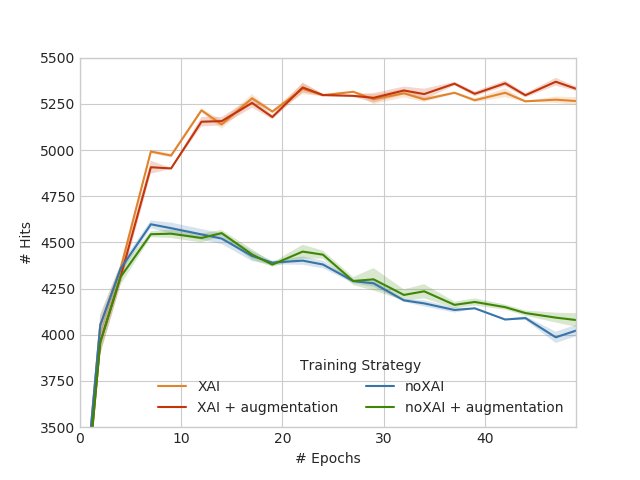} 
\caption{In this plot we present the number of hits, \ie the number of unseen MSCOCO images, among the 16K validation set, where the model is able to provide an accurate explanation for, among the correctly classified ones during training. A model explanation is defined in terms of a saliency map over the image pixels, and a better explanation is defined to be one that has a higher number of pointing game hits, \ie a higher number of image explanations overlapping with the annotation of the ground-truth class label. We can see that the noXAI model fits the dataset bias at training time, while the XAI model improves its explainability over time for validation data.}
\label{fig:localization_results}
\end{figure}


\begin{figure*}[!t]
    \centering 
    \includegraphics[width=0.993\linewidth,trim={0cm 0cm 0cm 0cm},clip]{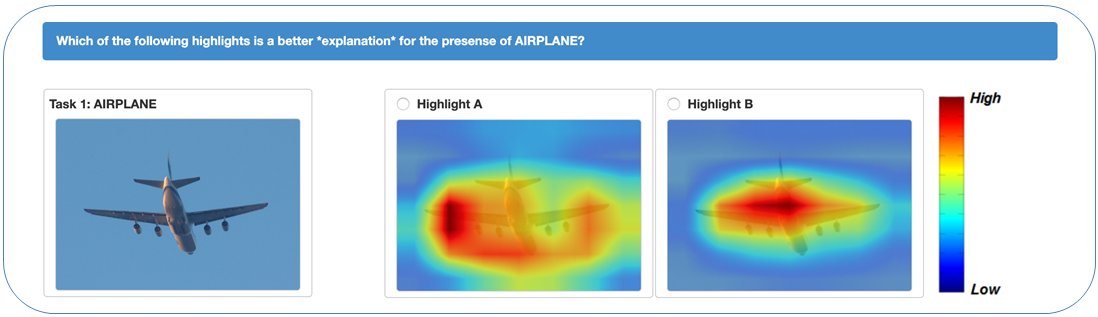}
\caption{A snapshot of the interface presented to Amazon Mechanical Turk users for a subtask. The interface asks the users to select the evidence (``highlight") they think is a better explanation for the presence of an object of a certain class, \eg airplane, together with the original image. The order in which the XAI and noXAI evidence maps are presented is randomized for every subtask image. A HIT consists of ten consecutive subtasks of the presented format and labeled by ten different crowd workers.}
\label{fig:interface}
\end{figure*}

\begin{figure*}[!t]
    \includegraphics[width=0.23\linewidth,trim={0cm 0.5cm 0cm 0cm},clip]{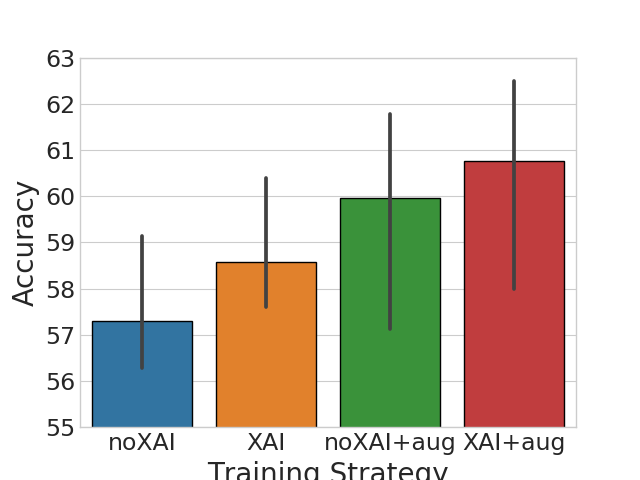}
    \hspace*{0.2em}
    \includegraphics[width=0.23\linewidth,trim={0cm 0.5cm 0cm 0cm},clip]{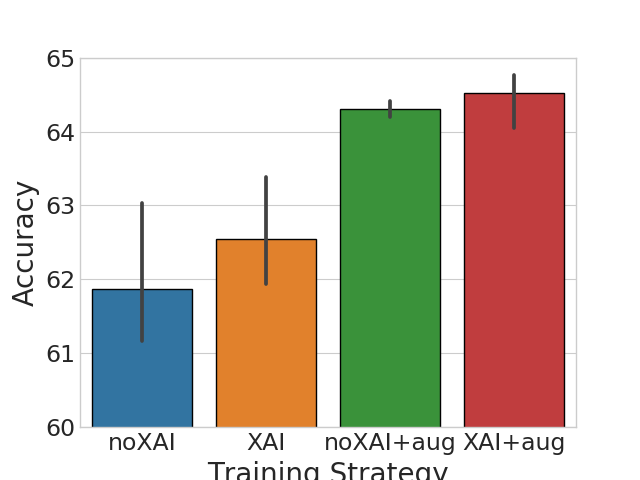}
    \hspace*{0.2em}
    \includegraphics[width=0.23\linewidth,trim={0cm 0.5cm 0cm 0cm},clip]{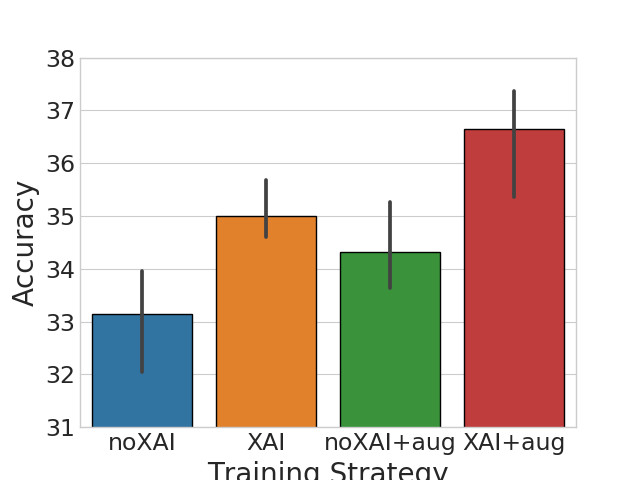}
    \hspace*{0.2em}
    \includegraphics[width=0.23\linewidth,trim={0cm 0.5cm 0cm 0cm},clip]{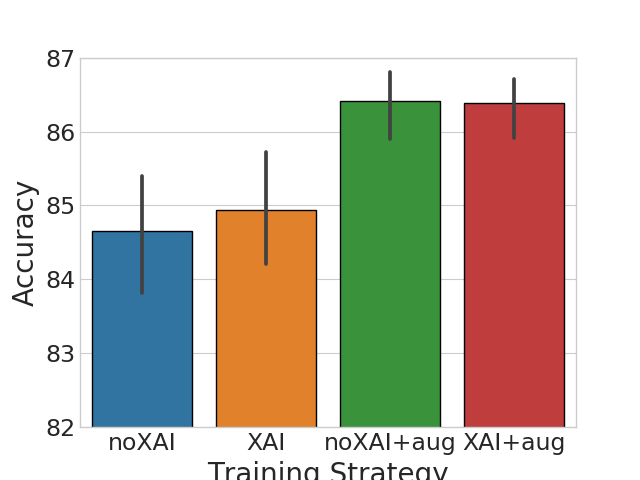}
\hspace*{4em}\textit{Graphics}\hspace*{9.5em}\textit{Clipart}\hspace*{9em}\textit{Infograph}\hspace*{9em} \textit{Painting}\\[0.05em]
   \includegraphics[width=0.23\linewidth,trim={0cm 0.5cm 0cm 0cm},clip]{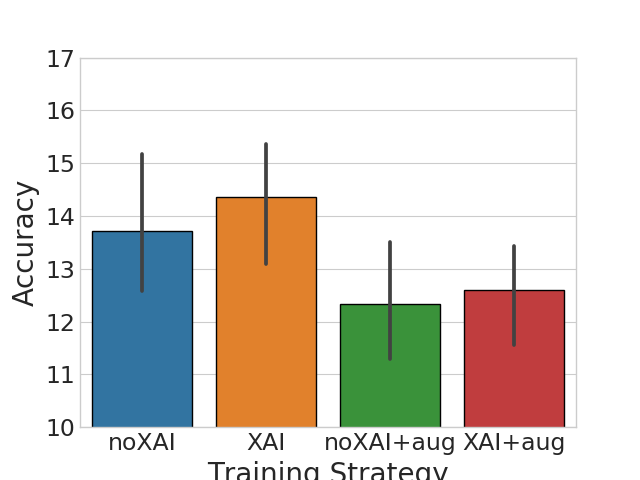}
    \hspace*{0.9em}
    \includegraphics[width=0.23\linewidth,trim={0cm 0.5cm 0cm 0cm},clip]{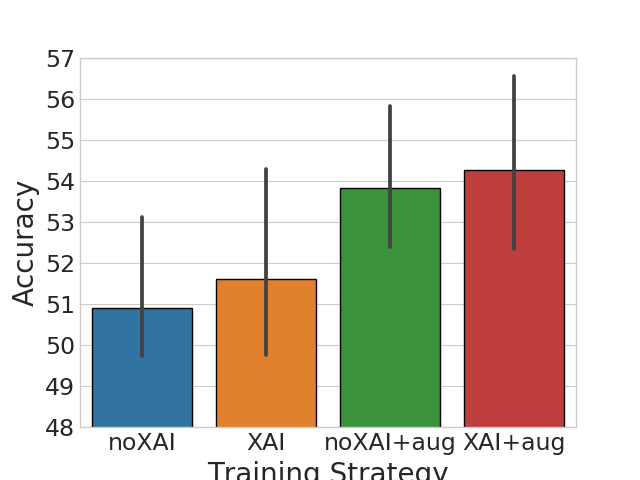}
    \hspace*{5em}
    \includegraphics[width=0.28\linewidth, ,trim={0cm 0.35cm 0cm 0cm},clip]{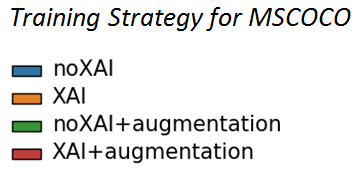}\\[0.05em]
    \hspace*{3em} \textit{Quickdraw} \hspace*{9em} \textit{Sketch}\hspace*{8em} \\[-0.05em]
    \vspace*{-0.6em}
    \caption{We present results for domain generalization on six \textit{unseen} target domains. The Graphics domain data is obtained from the Syn2Real dataset, and the other five domains are obtained from the DomainNet dataset. We note that training has been conducted on a single source: the MSCOCO dataset, and no data from any of the target domains is presented to the model at training time. Results are demonstrated for the common classes between the three datasets. For each training strategy we report the average test accuracy of the last four epochs for three trained models. The black bar demonstrates the minimum and maximum average over three trained models.}
\label{fig:results_COCO}
\end{figure*}

We explore how the noXAI and XAI models perform on the unseen validation set of MSCOCO. We investigate how localized the evidence is with respect to the ground-truth spatial annotations of MSCOCO. Figure~\ref{fig:noXAI_XAI} presents examples of saliency maps from the noXAI and XAI models after training is complete with comparable classification accuracy. Saliency is better localized over the object corresponding to the ground-truth class when the XAI strategy is adopted for training. We use the pointing game of~\cite{zhang2016top} to compute the number of hits; the number of correctly classified images where the peak saliency overlaps with the ground-truth spatial annotation. This is depicted in Figure~\ref{fig:localization_results} for the XAI and noXAI models over the training epochs to quantify explainability. We also demonstrate how data augmentation techniques of~\cite{volpi2019iccv} do not significantly affect localization of explanation maps. At epoch 50, the model accuracy for MSCOCO's validation set is: 61.87\% for the noXAI model and 62.04\% for the XAI averaged for three runs on the last four epochs. In essence, we are improving the explanation the model is providing for its classification result, assuming that a good explanation focuses more on the object corresponding to the ground-truth class, without hindering classification accuracy. The XAI model has learnt to rely less on context information, without hurting the performance. At epoch 50, the XAI model has 1321 images with better localization/explainability as per our automated metric. Next, we design a user study on these images to get an unbiased reflection of the human mental model of explainability.

\subsubsection{Human Judgment}

In Section~\ref{automated_metric} we assumed that a saliency map whose peak overlaps with the ground-truth spatial annotation of the object is a better explanation. We now design an unbiased crowdsourcing user study which asks users what they think is a better explanation for the presence of an object. 

\textbf{Annotation Tool Settings.} We use the Amazon Mechanical Turk (AMT) crowdsourcing marketplace to recruit crowd workers. We only accept AMT workers who had previously completed at least 1000 tasks (a.k.a HITs), and maintained
an approval rating of at least 98\%. We accept and compensate the work of all crowd workers who participated in our tasks.

\begin{figure*}[!t]
    \includegraphics[width=0.23\linewidth,trim={0cm 0.5cm 0cm 0cm},clip]{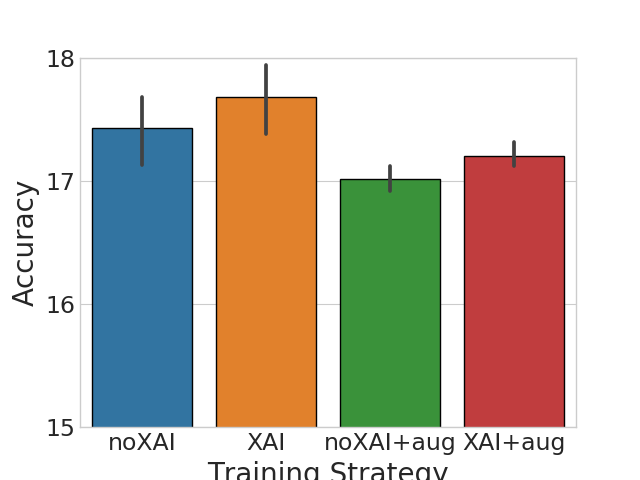}
    \hspace*{0.2em}
    \includegraphics[width=0.23\linewidth,trim={0cm 0.5cm 0cm 0cm},clip]{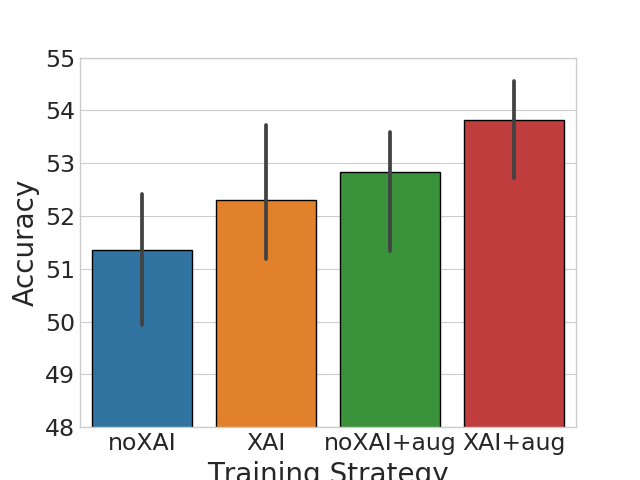}
    \hspace*{0.2em}
    \includegraphics[width=0.23\linewidth,trim={0cm 0.5cm 0cm 0cm},clip]{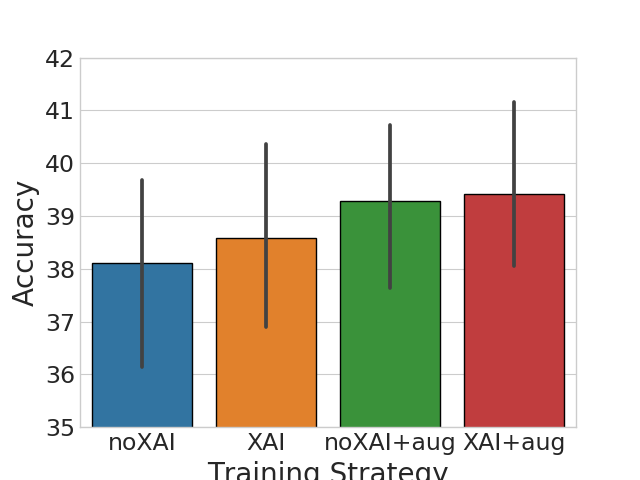}
    \hspace*{0.2em}
    \includegraphics[width=0.23\linewidth,trim={0cm 0.5cm 0cm 0cm},clip]{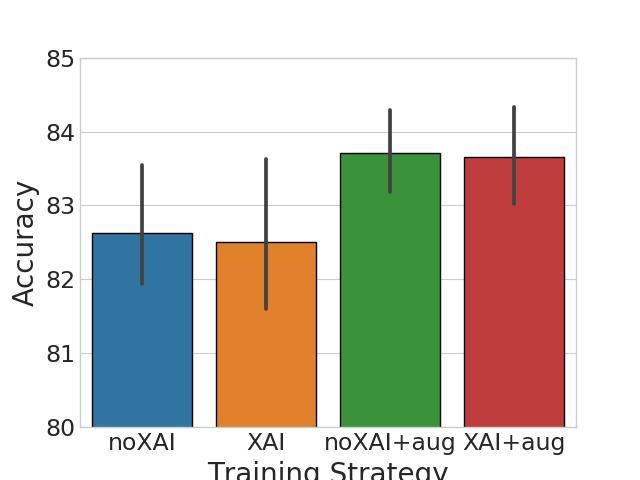}
\hspace*{4em}\textit{Graphics}\hspace*{9.5em}\textit{Clipart}\hspace*{9em}\textit{Infograph}\hspace*{9em} \textit{Painting}\\[0.05em]
   \includegraphics[width=0.23\linewidth,trim={0cm 0.5cm 0cm 0cm},clip]{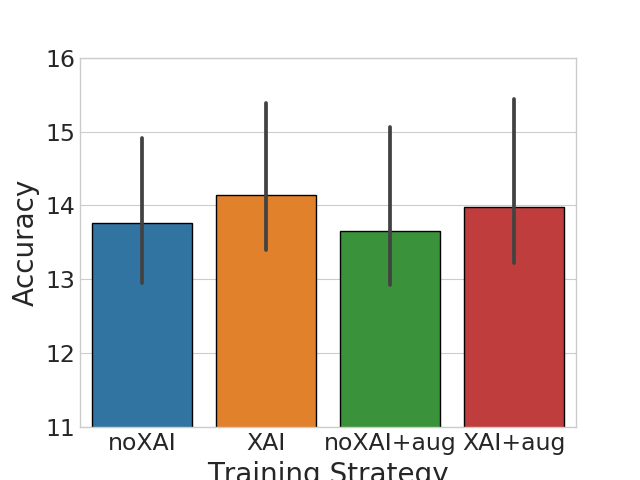}
    \hspace*{0.9em}
    \includegraphics[width=0.23\linewidth,trim={0cm 0.5cm 0cm 0cm},clip]{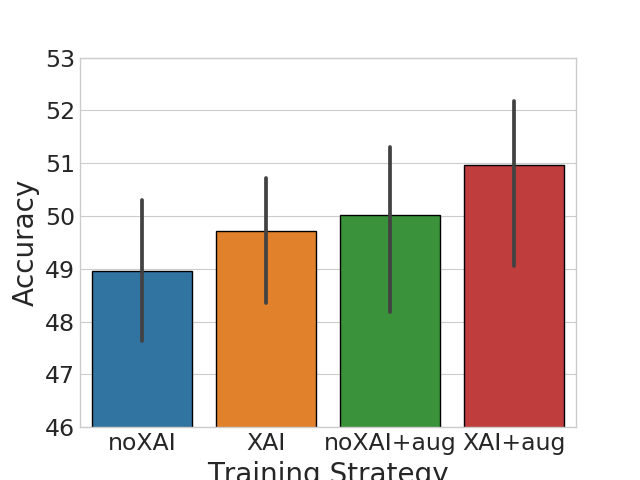}
    \hspace*{5em}
    \includegraphics[width=0.28\linewidth, ,trim={0cm 0.35cm 0cm 0cm},clip]{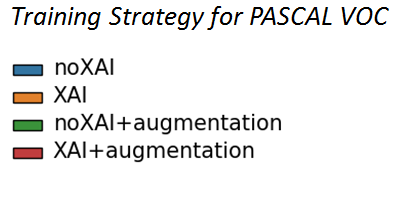}\\[0.05em]
    \hspace*{3em} \textit{Quickdraw} \hspace*{9em} \textit{Sketch}\hspace*{8em} \\[-0.05em]
    \vspace*{-0.6em}
    \caption{We present results for domain generalization on six \textit{unseen} target domains. The Graphics domain data is obtained from the Syn2Real dataset, and the other five domains are obtained from the DomainNet dataset. We note that training has been conducted on a single source: the PASCAL VOC dataset, and no data from any of the target domains is presented to the model at training time. Results are demonstrated for the common classes between the datasets. For each training strategy we report the average test accuracy of the last four epochs for three trained models. The black bar demonstrates the minimum and maximum average over three trained models.}
\label{fig:results_PASCAL}
\end{figure*}

\textbf{Crowdsourcing Task Details.} We collect annotations from
crowd workers for a task (HIT) that contains ten subtasks. Ten unique workers provide annotations for each task (annotating the ten subtasks). Each subtask presents the worker with one image and two evidence images, one is a saliency map generated from the model trained using our XAI strategy, and the second is a saliency map generated from the conventional noXAI training strategy. The worker is asked to select which of the evidence images (described to the user as ``highlights") is a better explanation for the presence of the object of interest. Figure~\ref{fig:interface} presents the sample interface presented to the worker for a particular subtask. We do not bias the users' definition of explainability by not providing any specific example/instruction images. We post all HITs simultaneously, for all 1321 images, while randomizing the presentation order of XAI and noXAI saliency maps in each subtask. We allot a maximum of ten minutes to complete each HIT and paid \$0.10 per HIT.

\textbf{Crowdsourcing Task Results.}
We processed the 13,210 (1,321 images * 10 votes per image) crowdsourced results to assign a winning label, ``XAIwinner” or ``noXAIwinner”, for each image. 190 unique crowd workers contributed to the 10 votes per HIT. It took each crowd worker an average of 137 seconds to complete a HIT consisting of ten voting subtasks. We used majority voting to combine the ten crowd-collected selections into a single vote for each image. For 225 out of the 1,320 images, there was no clear winner as the XAI evidence and noXAI evidence were each chosen by 5 out of 10 workers. For the remaining 1,096 images, XAI evidence won for 887 of the images (67\% of the whole image population and 80\% of the images with a winner choice).

\subsection{Improved Domain Generalization}


In this section, we explore whether training a model to be explainable helps generalizing to unseen domains. We consider the challenging problem of domain generalization from a \textit{single-source} dataset. This means that the target domains are completely unseen during training. We use MSCOCO as our single source and utilize its ground-truth spatial annotations at training time to enforce a more explainable model. We then test noXAI and XAI models on domains that are unseen at training time: graphics, clipart, infograph, painting, quickdraw, and sketch. 

 Figure~\ref{fig:results_COCO} presents the classification accuracy on unseen domains of XAI and no XAI models, with and without data augmentation \cite{volpi2019iccv}. The presented results are the average of the last four epochs, and every model is trained three times. The XAI model consistently results in improved domain generalization over the six target domains tested. The XAI training strategy achieves a relative 2.56\% accuracy improvement averaged over all target domains. Figure~\ref{fig:results_COCO} also demonstrates the complementarity of our proposed XAI training strategy to that of Volpi and Murino~\cite{volpi2019iccv}. Figure~\ref{fig:results_PASCAL} presents a similar trend when the PASCAL VOC is used as the single-source dataset for enforcing explainability, and unseen target domains are used for testing from DomainNet and Syn2Real.

By guiding the network to periodically focus on regions that contain the object of a ground truth class, we are able to train models that are more explainable in the sense that they classify based on the correct evidence, generalize better to unseen target domains, and suffer no degradation in performance on the original domain.





\begin{table*}[t!]
\centering

\begin{tabular}{l|c|c|c|c} 
\hline
\textbf{Domain} & \textbf{After block1} & \textbf{After block2}   & \textbf{After block3}  & \textbf{After block4}  \\
\hline
Graphics & \bf{58.67} {\small }  & 58.02 {\small }&57.09 {\small } & 58.56 {\small }   \\
Clipart & \bf{63.90} {\small }& 63.21 {\small }& 62.55 {\small } & 62.59 {\small } \\
Infograph & 32.74 {\small } & 32.68 {\small }  & 32.91 {\small } & \bf{35.16} {\small }    \\
Painting & 85.30 {\small } & \bf{85.31} {\small } & 85.10 {\small } & 84.98 {\small }    \\
Quickdraw & 14.29 {\small }  & 13.85 {\small } & 13.13 {\small } & \bf{14.49} {\small } \\
Sketch & 51.74 {\small }  & 51.40 {\small } &51.32 {\small }  & \bf{51.76} {\small }\\
\hline
\end{tabular}
\vspace*{0.5em}
\caption{Test accuracies on the six unseen domains from DomainNet and Syn2Real, where XAI training was applied at different locations (After block1, After block2, After block3, and After block4) in the ResNet-50 architecture.  }
\label{table:ablation_layer}
\end{table*}

\begin{table*}[t]
\centering
\begin{tabular}{l|c|c|c} 
\hline
\textbf{Domain} 
& \textbf{XAI ($freq = 5$)}  & \textbf{XAI ($freq = 10$)} & \textbf{XAI ($freq = 15$)} \\
\hline
Graphics  
& \bf{58.56} & 58.06  & 58.08    \\
Clipart  
& \bf{62.59} & 61.69  & 62.22  \\
Infograph  
& \bf{35.16}   & 33.91  & 34.07   \\
Painting  
& \bf{84.98}  & 84.86  & 84.96    \\
Quickdraw  
& \bf{14.49}  & 13.96  & 14.13  \\
Sketch  
&\bf{51.76}  &51.04   & 51.42\\
\hline
\end{tabular}
\vspace*{0.5em}
\caption{Test accuracies on the six unseen domains from DomainNet and Syn2Real, where XAI training was applied at $freq=5, 10, 15$, \ie every 5, 10, or 15 training epochs.}
\label{table:ablation_freq}
\end{table*}

\subsection{Where and When to apply XAI training}
\label{wherewhen}

In this section we perform two ablation studies on MSCOCO to explore where and when XAI training could be applied to maximize performance gains.

\textbf{Where to apply XAI training?} In this section we explore where in a network architecture is best for applying the XAI training. This is where we would be computing the saliency map for the images, computing a mask, and applying the mask on forward activations. We apply XAI training after every block of the ResNet-50 architecture and present the results in Table~\ref{table:ablation_layer}. The majority of the domains had highest accuracy when XAI training was applied after the last block, however, we found that accuracies are comparable regardless to where XAI training was applied.

\textbf{When to apply XAI training?} 
We apply XAI training using $freq=5,10,15$ epochs in Table~\ref{table:ablation_freq}. Frequencies greater than $1$ obtain a comparable performance, with best performance achieved at $freq=5$. Applying XAI every iteration would involve a computational overhead and would completely disregard context hindering performance on the source and target domains.

\section*{Conclusions}


In this work, we make object classification models possess the valuable property of explainability exposing their internal decision process in a human-interpretable way. We do so by forcing the model to focus on the evidence, in the form of a saliency map, of corresponding objects that have been labeled by humans at training time, and observe similar patterns at test time. We demonstrate that the model has learnt to classify objects by looking at the objects themselves and not on the surrounding context. We also demonstrate that this leads the model trained using this strategy to better generalize to other domains. While our approach leads to more explainable classification models and better generalization to unseen domains, it has no associated accuracy degradation on the original domain and no added test-time complexity.

{\small
\bibliographystyle{ieee_fullname}
\bibliography{egbib}
}

\end{document}